\documentclass[runningheads]{llncs}
\usepackage{graphicx}
\usepackage{subcaption}
\usepackage{comment}
\usepackage{amsmath,amssymb} 
\usepackage{color}

\DeclareMathOperator*{\argmin}{arg\,min}

\graphicspath{{./figure/}{./}}

\begin{document}
\pagestyle{headings}
\mainmatter
\def\ECCVSubNumber{3441}  

\title{SMART: Skeletal Motion Action Recognition aTtack} 

\titlerunning{SMART}
%
\author{He Wang\inst{1} \and Fexiang He\inst{1} \and Zhexi Peng\inst{2} \and Yongliang Yang\inst{3} \and Tianjia Shao\inst{4} \and Kun Zhou\inst{4} \and David Hogg\inst{1}}
\authorrunning{F. Author et al.}
%
\institute{University of Leeds, UK \and Beijing University of Posts and Telecommunications, China \and University of Bath, UK \and Zhejiang University, China}
\maketitle

\begin{abstract}
Adversarial attack has inspired great interest in computer vision, by showing that classification-based solutions are prone to imperceptible attack in many tasks. In this paper, we propose a method, \emph{SMART}, to attack action recognizers which rely on 3D skeletal motions. Our method involves an innovative perceptual loss which ensures the imperceptibility of the attack. Empirical studies demonstrate that SMART is effective in both white-box and black-box scenarios. Its \textbf{generalizability} is evidenced on a variety of action recognizers and datasets. Its \textbf{versatility} is shown in different attacking strategies. Its \textbf{deceitfulness} is proven in extensive perceptual studies. Finally, SMART shows that adversarial attack on 3D skeletal motions, one type of time-series data, is significantly different from traditional adversarial attack problems.
\end{abstract}

\section{Introduction}

Adversarial attack has invoked a new wave of research interest recently. On the one hand, it shows that deep learning models, as powerful as they are, are vulnerable to attack, leading to security and safety concerns \cite{Szegedy:2014:SRA}; on the other hand, it has been proven to be useful in improving the robustness of existing models \cite{Liao_2018_CVPR}. Starting from object recognition, the list of target tasks for adversarial attack has been rapidly expanding, now including face recognition \cite{Sharif16AdvML}, point clouds \cite{Xiang:2019:PointAdv}, and 3D meshes \cite{Xiao:2019:MeshAdv}. While adversarial attack on static data (images, geometries, etc.) has been explored, its effectiveness on time-series has only been attempted under general settings \cite{DBLP:journals/corr/abs-1902-10755}. In computer vision, video-based attack has been attempted in attacking recognition tasks \cite{DBLP:journals/corr/abs-1803-02536}. In this paper, we would like to attack another type of time-series data: 3D skeletal motions, for action recognition tasks.

Skeletal motion has been widely used in action recognition \cite{Du:2015:HRN}. It can greatly improve recognition accuracy by mitigating issues such as lighting, occlusion and posture ambiguity. In this paper we show that 3D skeletal motions are also vulnerable to adversarial attack and it can thus cause serious concerns. Adversarial attack on 3D skeletal motion faces two unique and related challenges which are significantly different from other attack problems: low redundancy and perceptual sensitivity. When attacking images/videos, it is possible to perturb some pixels without causing too much visual distortion. This largely depends on the redundancy in image space \cite{Tramr2017TheSO}.  Unlike images, which have thousands of Degrees of Freedom (DoFs), a motion frame (or a pose) in 3D skeletal motions is usually parameterized by fewer than 100 DoFs (in our experiments, we use 25 joints, equivalent to 25*3=75 DoFs). This not only restricts the space of possible attacks, but also has severe consequences on the imperceptibility of the adversarial examples: a small perturbation on a single joint can be easily noticed. Furthermore, coordinated perturbations on multiple joints in only one frame can hardly work either, because in the temporal domain, similar constraints apply. Any sparsity-based perturbation (on single joints or individual frames) will greatly affect the dynamics (causing jittering or bone-length violations) and will be very obvious to an observer. 

We propose an adversarial attack method, SMART, based on an optimization framework that explicitly considers motion dynamics and skeletal structures. The optimization finds perturbations by balancing between classification goals and perceptual distortions, formulated as classification loss and perceptual loss. Varying the classification loss leads to different attacking strategies. The new perceptual loss fully utilizes the dynamics of the motions and bone structures. We empirically show that SMART is effective in both white-box and black-box settings, on several state-of-the-art models, across a variety of datasets.

Formally, our contributions include:
\begin{itemize}
    \item A novel perceptual loss function for adversarial attack on action recognition based on 3D skeletal motions. The new perceptual loss captures the perceptual realism and fully exploits the motion dynamics.

    \item Empirical evidence that 3D skeletal motions are vulnerable to attack under multiple settings and attacking strategies, by extensive experiments and user studies.
    
    \item Insights into the role of dynamics in the imperceptibility of adversarial attack based on comprehensive perceptual studies. This result differs significantly from widely accepted approaches, which use $l$-norm with $\epsilon$-ball constraints on static data such as images.
\end{itemize}

\section{Related Work}
\subsection{Skeleton-based Action Recognition}
Action recognition is crucial for many important applications, namely visual surveillance, human-robot interaction and entertainment. Recent advances in 3D sensing and pose estimation motivate the use of clean skeleton data to robustly classify human actions, overcoming the biases from raw RGB video due to body occlusion, scattered background, lighting variation, etc. Unlike conventional approaches that are limited to handcrafted skeletal features~\cite{Vemulapalli:2014:HAR,Fernando:2015:MVE,Devanne:2015:HAR}, recent methods taking the advantage of trained features from deep learning have gained the state-of-the-art performance. According to the representation of the skeleton data used for training, deep learning based methods can be classified into three categories, including sequence-based methods, image-based methods, and graph-based methods, respectively.

Sequence-based methods represent skeleton data as a chronological sequence of postures, each of which consists of the coordinates of all the joints. Then RNN-based architecture is employed to perform the classification~\cite{Du:2015:HRN,Shahroudy:2016:NTU,Liu:2016:STL,Song:2017:ESA,Zhang:2019:VAN}. Image-based methods represent skeletal motion as a pseudo-image, which is a 2D tensor where one dimension corresponds to time, and the other dimension stacks all joints from a single skeleton. Such representation enables CNN-based image classification to be applied in the action recognition context~\cite{Liu:2017:ESV,Ke:2017:NRS}. Different from the previous two categories that mainly rely on skeleton geometry represented by the joint coordinates, graph-based methods utilize graph representations to naturally consider the skeleton topology (i.e. joint connectivity) encoded by bones that connect neighboring joints. Graph neural networks (GNN) are used to recognize the actions~\cite{Yan:2018:STG,Tang:2018:DPR,Shi:2019:SBA}. Based on the code released by the original authors, we perform adversarial attacks on two most representative categories (i.e. RNN- and GNN-based), demonstrating the vulnerability of different types of neural networks.

\subsection{Adversarial Attacks}
Despite their significance in enhancing vision-based tasks such as classification and recognition, deep neural networks are vulnerable to carefully crafted adversarial attacks as firstly pointed out in~\cite{Szegedy:2014:SRA}. In other words, delicately designed neural networks with high performance can be easily fooled by an unnoticeable perturbation on the original data. With the above concern raised, researchers have extensively investigated adversarial attacks on different data types, including 2D images~\cite{Goodfellow:2014:explaining,Papernot:2015:LDL,Dezfooli:2016:DeepFool,Xiao:2018:GAE,Xiao:2018:Spatially}, videos~\cite{Wei:2018:SparseAP,Wang:2018:LDV}, 3D shapes~\cite{Liu:2019:BPN,Zeng:2019:AAB,Xiao:2019:MeshAdv,Xiang:2019:PointAdv}, physical objects~\cite{Kurakin:2016:AEP,Athalye:2017:SRA,Evtimov:2017:RPW}, while little attention has been paid on 3D skeletal motions.

Adversarial attacks in the context of action recognition is much less explored. Inkawhich et al.~\cite{Inkawhich:2018:AAO} perform adversarial attacks on optical-flow based action recognition classifiers, which is mainly inspired by image-based attacks and differs from our work in terms of the input data.

In recent contemporaneous work to our own~\cite{Liu:2019:AAS} on arXiv, adversarial attack is applied to a GNN network for action classification on skeletal motions~\cite{Yan:2018:STG}. The loss function used for the attack minimizes the joint position deviation and acceleration, where the attack however is noticeable.  In our work, we demonstrate better results using a perceptual loss that minimizes the motion derivative deviation relative to the original skeletal motion, thereby preserving the motion dynamics intrinsic to actions. This is crucial in attacking highly dynamic motions such as running and jumping. We also perform a perceptual study to validate the imperceptibility of the perturbed skeletal motions and the effectiveness of our choice of perceptual loss.

We demonstrate successful attacks on a range of network architectures, including RNN and GNN based methods, on three datasets. Finally, we present results of three different attacking strategies, including the novel objective of placing the correct action beneath the first n actions in a ranked classification, for a given n.


\section{Methodology}
SMART is formulated as an optimization problem, where the minimizer is an adversarial example, for a given motion, that minimizes the perceptual distortion while fooling the target classifier. The optimization has three variants constructed for three different attacking strategies: \textit{Anything-but Attack}, \textit{Anything-but-N Attack} and \textit{Specified Attack}. They are used in \textit{white-box} and \textit{black-box} scenarios.

\subsection{Optimization for Attack}
In an action recognition task, given a motion $q$ = \{$q_0$, $q_1$, ... ,$q_t$\}, where $q_t$ is the frame at time $t$ and consists of stacked 3D joint locations, a trained classifier $\Phi$ can predict its class label $y_q$ = $C(\Phi(q)$), where $\Phi$ is namely a deep neural network and $\Phi(q)$ is the predicted distribution over class labels. $C$ is usually a \textit{softmax} function and $y_q$ is the predicted label. We aim to find a perturbed example, $\hat{q}$, for $q$, such as $y_q \neq y_{\hat{q}}$. 

Without any constraints, it is trivial to find $\hat{q}$. So normally, it requires that the difference between $q$ and $\hat{q}$ is not perceptible. This can be formulated into a generic optimization problem:
\begin{equation}
    \argmin_{\hat{q}}~L = \argmin_{\hat{q}}~w L_c(q, \hat{q}) + (1-w)L_p(q, \hat{q})
\end{equation}
where $L_c$ and $L_p$ are a classification loss and a perceptual loss respectively and $w$ is a weight. We use $w=0.4$. Intuitively, there are two forces governing $\hat{q}$. $L_c$ is the classification loss where we can design different attacking strategies. $L_p$ is the perceptual loss which dictates that $\hat{q}$ should be visually indistinguishable from $q$. To optimize for $\hat{q}$, we have only one mild assumption: we can compute the gradient: $\frac{\partial L}{\partial \hat{q}}$. This way, we can compute $\hat{q}$ iteratively by $\hat{q}_{t+1}$ = $\hat{q}_{t}$ + $\epsilon f(\frac{\partial L}{\partial \hat{q}_t}, \hat{q}_t)$ where $\hat{q}_t$ is $\hat{q}$ at step $t$, $f$ computes the updates and $\epsilon$ is the learning rate. We set $\hat{q}_0$ = $q$ and use Adam \cite{Diederik_2014} for $f$.

\subsection{Perceptual Loss}
\label{sec_pl}
Imperceptibility is a hard constraint in adversarial attacks. It requires that human cannot distinguish easily between the adversarial examples and real data. Many existing approaches on images and videos achieve imperceptibility by computing the image-wise or frame-wise minimal changes, measured by a certain type of $l$ norm, e.g. $l_1$, $l_2$ or $l_{\infty}$. However, it would not work for motions because they do not consider dynamics. 

To fully represent the dynamics of a motion, we need the derivatives from \textit{zero-order} (joint location), \textit{first-order} (joint velocity)  up to \textit{nth-order}. One common approximation is to use first n terms, e.g. up to the second-order. When it comes to imperceptibility on motions, the perceived motion naturalness is vital and not all derivatives are at the same level of importance \cite{Wang_STRNN_2019}. Inspired by the work in character animation \cite{Wang_STRNN_2019,wang_harmonic_2013}, we propose a new perceptual loss:
\begin{eqnarray}
    L_p(q, \hat{q}) = \alpha l_{dyn} + (1-\alpha)l_{bl} \\
    l_{bl} = ||Bl(q) - Bl(\hat{q})||_2^2 = \frac{1}{M}\sum_{i=1}^{M} ||Bl(q_i) - Bl(\hat{q}_i)||_2^2\\
    l_{dyn} = \sum_{n=0}^{\infty} \beta_n||\gamma(q^n - \hat{q}^n)||_2^2~~\textrm{where} \sum_{n=0}^{\infty} \beta_n = 1
\end{eqnarray}
where $\alpha$ is a weight and set to 0.3 for our experiments. $l_{bl}$ penalizes any bone length deviations in every frame where $M$ is the total frame number. $Bl(q_i) \in$ $\mathbb{R}^{24\times1}$ is the bone length vector of frame $q_i$. Theoretically, the bone lengths do not change over time. However, due to motion capture errors, they do vary in different frames even in the original motions. This is why $l_{bl}$ is designed to be a frame-wise bone length loss term. 

$l_{dyn}$ is the dynamics loss. We use a strategy called \textit{derivative matching}. It is a weighted (by $\beta_n$) sum of $l_2$ distance between $q^n$ and $\hat{q}^n$, where $q^n$ and $\hat{q}^n$ are the $nth$-order derivatives and can be computed by forward differencing. $\gamma$ is a 75$\times M$-dimensional vector of weights (75 DoFs per frame for $M$ frames). Although $n$ goes up to infinity, in practice, we explored up to $n=4$, which includes joint position, velocity, acceleration, jerk and snap. After exhaustive experiments, we found that a good compromise is to set $\beta_0 = 0.6$, $\beta_2 = 0.4$ and the rest to 0. Matching the 2$nd$-order profiles of two motions is critical. For skeletal motions, small location deviations can still generate large acceleration differences, resulting in two distinctive motions. More often, it generates severe jittering and thus totally unnatural motions. An alternative way of regulating the dynamics is to purely smooth the motion, by e.g. minimizing the acceleration. But it damps highly dynamic motions such as jumping~\cite{Wang_STRNN_2019}. Also, considering more derivatives above $n=4$ makes the optimization harder to solve and over-weighs their benefits.

Finally, we fix the values of $\gamma$s. Based on our preliminary studies, we found that the perceived motion naturalness is not affected by all joints equally. The jittering on the torso has a higher impact. So we use higher weights on the spinal joints. For all our experiments, the skeleton has 25 joints and 24 bones. We use 0.04 for the DoFs of spinal joints and 0.02 for the rest.

\subsection{White-box Attack}
\label{sec:whitebox}
With the perceptual loss fixed, varying the formulation of the classification loss allows us to form different attacking strategies. We present three strategies.

{\bf Anything-but Attack (AB)}. Anything-but Attack aims to fool the classifier so that $y_q \neq y_{\hat{q}}$. This can be achieved by maximizing the \textit{cross entropy} between $\Phi(q)$ and $\Phi(\hat{q})$:
\begin{equation}
    L_c(q, \hat{q}) = -cross\_entropy(\Phi(q), \Phi(\hat{q}))
\end{equation}
Comparatively, AB is the easiest optimization problem among the three strategies. $\Phi(\hat{q})$ could peak on any class label but the ground-truth or even become just flat.

{\bf Anything-but-N Attack (ABN)}. Anything-but-N Attack is a generalization of AB. It aims to confuse the classifier so that it has similar confidence levels in multiple classes. ABN is more suitable to confuse classifiers which rely on top N accuracy. In addition, we found that it performs better in black-box attacks by transferability, which will be detailed in experiments. 

Instead of simply using multiple AB losses for the top N classes, we propose an easier loss function, maximizing the entropy of the predicted distribution of $\hat{q}$:
\begin{eqnarray}
    L_c(q, \hat{q}) = -Entropy(\Phi(\hat{q})) \\ 
    y_q \not\in TopN(\Phi(\hat{q})) \nonumber
\end{eqnarray}
where $TopN$ is the set of the top n class labels in the predictive distribution $\Phi(\hat{q})$. By minimizing $L_c$, we actually maximize the entropy of $\Phi(\hat{q})$, i.e. forcing it to be flat over all class labels and thus reduce the confidence of the classifier over any particular class. We stop the optimization once the ground-truth label falls beyond the top n classes. ABN is a harder optimization problem than AB because it needs the predictive distribution to be as flat as possible.

\subsubsection{Specified Attack (SA)}
Different from AB and ABN, sometimes it is useful to fool the classifier with a specific class label. Given a fake label $y_{\hat{q}}$, we can compute its corresponding class label distribution $\Phi_{\hat{q}}$ and minimize the cross entropy:
\begin{equation}
    L_c(q, \hat{q}) = cross\_entropy(\Phi(\hat{q}), \Phi_{\hat{q}})
\end{equation}
This is the most difficult scenario because it is highly related to the similarity between the source and target labels. Turning a `clapping over the head' motion into a `raising two hands' could be relatively achievable and cause minimal visual changes; while turning a `running' motion into a `squat' motion without noticeable visual changes is much harder.

\subsection{Black-box Attack}
Black-box attack assumes that only very little information about the target classifier is known. Under such circumstances, we use \textit{attack-via-transferability} \cite{Tramr2017TheSO}. It begins with training a surrogate classifier. Then adversarial examples are computed by attacking the surrogate classifier. Finally, the adversarial examples can be used to attack the target classifier. In this paper, we do not construct our own surrogate model. Instead, we use an existing classifier as our surrogate classifier to attack the others. In experiments, we attack several state-of-the-art models. To test the transferability and generalizability of our method, we use every model in turns as the surrogate model and attack the others.

\section{Experimental Results}
We first introduce the datasets (Section~\ref{exp_data}) and models (Section~\ref{exp_model}) for our experiments. Then we present our white-box (Section~\ref{exp_white}) and black-box (Section~\ref{exp_black}) attack results. Finally, we present our perceptual studies on imperceptibility (Section~\ref{exp_study}). Since we attack multiple models on multiple datasets, we first use the source code shared by the authors if available or implement the models ourselves. Then we train them strictly following the protocols in their papers. Next, we test the models and collect the data samples that the trained classifiers can successfully recognize, to create our adversarial attack datasets. Finally, we compute the adversarial samples using different attacking strategies.

\subsection{Datasets}
\label{exp_data}
When choosing datasets, our criteria are: 1. It needs to be widely used and contain 3D skeletal motions. 2. The motion quality needs to be as high as possible because it is tightly related to our perceptual study. Finally, we chose 3 benchmark datasets:

\textbf{HDM05 dataset~\cite{cg-2007-2}} is a 3D motion database captured with an Mocap system. It contains 2337 sequences for 130 actions performed by 5 non-professional actors. The 3D joint locations of the subjects are provided in each frame.

\textbf{Berkeley MHAD dataset~\cite{6474999}} is captured using a multi-modal acquisition system.  It consists of 11 actions performed by 12 subjects, where 5 repetitions are performed for each action, resulting in 659 sequences. In each frame the 3D joint positions are extracted based on the 3D marker trajectories.

\textbf{NTU RGB+D dataset~\cite{Shahroudy_2016_NTURGBD}} is captured by Kinect v2 and is currently one of the largest publicly available datasets for 3D action recognition. It is composed of more than 56,000 action sequences. A total of 60 action classes are performed by 40 subjects. The videos are captured from 80 distinct viewpoints. The 3D coordinates of joints are provided by the Kinect data. Due to the huge number of samples and the large intra-class and viewpoint variations, the NTU RGB+D dataset is very challenging and is highly suitable to validate the effectiveness and generalizability of our approach.

\subsection{Target Models}
\label{exp_model}
We selected 5 state-of-the-art methods: HRNN~\cite{Du_CVPR_2015}, ST-GCN~\cite{Yan_AAAI_2018}, AS-GCN~\cite{Li_2019_CVPR}, DGNN~\cite{Shi:2019:SBA} and 2s-AGCN~\cite{Shi_2019_CVPR}. They include both RNN- and GNN-based models. We implemented HRNN following the paper and used the code shared online for the rest four methods.

\begin{figure}[tb]
  \includegraphics[width=0.155\textwidth]{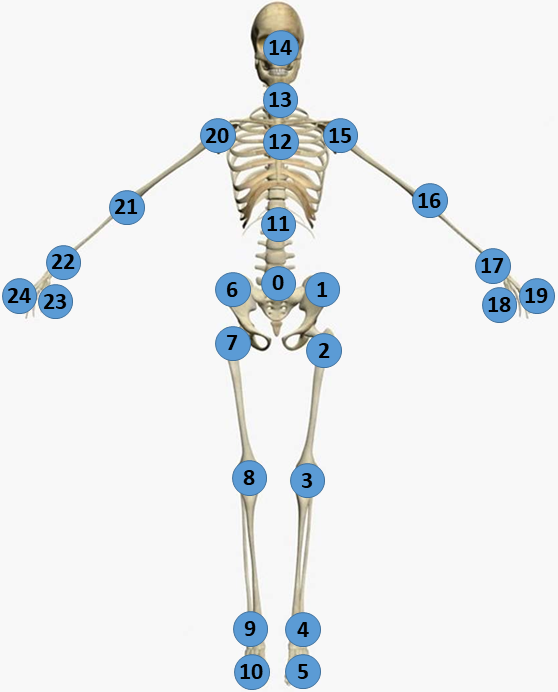}
  \includegraphics[width=0.84\linewidth]{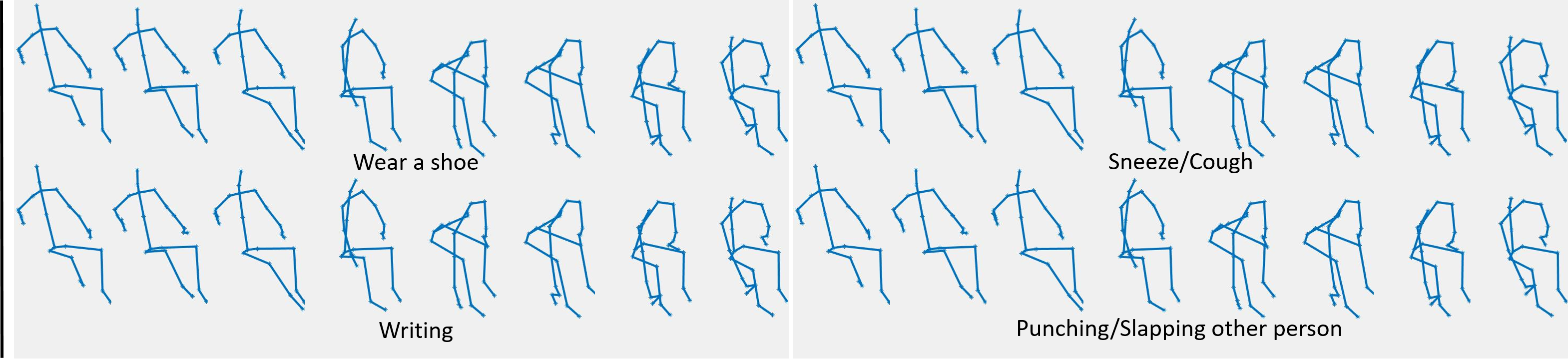}
  \caption{Left: Joint labels. Right: AS-GCN on NTU. Top: Original and AB. Bottom: AB5 and SA. `Wear a shoe' is the ground-truth label.}
  \label{fig:whitebox}
\end{figure}

We also followed their protocols in data pre-processing. Specifically, we preprocess the HDM05 dataset and Berkeley MHAD dataset as in~\cite{Du_CVPR_2015}, and the NTU RGB+D dataset as in~\cite{Shi_2019_CVPR}. Their respective class numbers are 65, 11 and 60. We also map different skeletons to a standard 25-joint skeleton as in \cite{Wang_STRNN_2019} (Figure \ref{fig:whitebox} Left). Please refer to relevant papers for details.


The five target models require different inputs, but they can be easily unified. HRNN, ST-GCN and AS-GCN all take joint positions in each frame as input. DGNN and 2s-AGCN require joint positions and bones.  Although bones are taken as an independent input, they can be computed from joint positions.  So we add another input layer before the original model to compute bones from joint positions. As this layer is only for computing bones and introduces no new variables, it does not change the behaviours of the original models.


\subsection{White-box Attack}
\label{exp_white}
In this section, we qualitatively and quantitatively evaluate the performance of SMART on the aforementioned three datasets. We use a learning rate between 0.005 and 0.0005 and a maximum of 300 iterations. The setting for AB and ABN is straightforward. In SA, the number of experiments needed will be prohibitively large if we attack every motion with every other label but the ground-truth. Instead, we randomly select fake labels to attack. Since the number of motions attacked is large, the results are adequately representative. Note that this is a very strict test as most of the motions are rather distinctive.

To perform the attack, we first train the target models using the settings in the original papers to ensure similar training results. We then test the models with the testing dataset and gather the motions that can be successfully recognized. Lastly, we attack these motions and record the success rates. For simplicity, we only show representative results in the paper. For more comprehensive results, please refer to the supplementary materials and video.

\subsubsection{Attack Results.}


First, we shows one motion in NTU attacked using three strategies in Figure \ref{fig:whitebox} Right. `Wear a shoe' is attacked to `Sneeze/Cough', `Writing' and `punching/Slapping other person' respectively in three attacking strategies. Although the semantics of the action labels are distinctive, the perturbation is hardly noticeable. We also show the quantitative results of AB in Table \ref{table_attack} Left. High success rates are universally achieved across different datasets and different target models, demonstrating the generalizability of SMART. For adversarial attack, it is not surprising if the before-attack and after-attack labels are semantically similar, e.g. from drinking water to eating. In SMART, a variety of examples are found where the after-attack labels are significantly different from the original ones. Due to the space limit, we leave all the details in the supplementary video and materials and only give a couple of examples here.  In HDM05, high confusion was found between turn\_L (turn left) and walk\_rightRC (walk sideways, to the right, feet cross over alternately front/back) for HRNN (Figure 16, supp\_hdm05.pdf). Similarly, in NTU, high confusion was found between standing\_up (from sitting position) and wear\_a\_shoe for 2SAGCN (Figure 20, supp\_ntu.pdf). These labels have completely different semantics and involve different body parts and motion patterns. Moreover, this kind of confusion is observed across all datasets and models. 


\begin{table}[tb]
	\centering
	\resizebox{0.95\textwidth}{!}{
		\begin{tabular}{c|ccc|ccc|ccc}		
			\hline
			Model/Data & HDM05 & MHAD  & NTU & HDM05 & MHAD  & NTU & HDM05 & MHAD  & NTU \\ \hline
			HRNN  & 100   & 100 & 99.56 & 100/100 & 100/100 & 99.84/99.62 & 67.19  & 57.41 & 49.17      \\
			ST-GCN & 99.57   & 99.96 & 100 & 93.30/90.28 & 76.86/70.5 & 95.86/91.32 & 74.95 & 66.93 & 100      \\
			AS-GCN & 99.36   & 92.84 & 97.43 & 91.46/82.83 & 42.07/22.34 & 91.18/82.47 & 64.62 & 40.18 & 99.48    \\
			DGNN & 96.09   & 94.46 & 92.51 & 93.55/86.32 & 87.54/74.27 & 98.73/97.62 & 97.26 & 96.13 & 99.99 \\
			2s-AGCN & 99.18   & 95.97 & 100 & 83.40/75.2 & 55.9/32.08 & 100/100 &  96.72 & 97.53 & 100  \\
			\hline
			mean & 98.84 & 96.65 & 97.9 & 92.34/86.93 & 72.47/59.84 & 97.12/94.21 & 80.15 & 71.64 & 89.73     \\ \hline
		\end{tabular}
    }
	\caption{Success rate. Left: Anything-but (AB) Attack. Mid: Anything-but-N Attack. The results are AB3/AB5 when n = 3 (AB3) and 5 (AB5). Right: Specified Attack (SA).}
	\label{table_attack}
\end{table}

We show the ABN results in Table~\ref{table_attack} Mid, in two variations: AB3 and AB5, as a generalization of AB. They are good for attacking classifiers based on top N accuracy. Although the overall performance is still good, the success rates are relatively lower compared with AB. It verifies our qualitative analysis in Section \ref{sec:whitebox}. ABN is harder than AB. Also AB5 is harder than AB3. In addition, the results on MHAD is not as good as the other two. This is because there are only 11 classes as opposed to 65 and 60 in the other two. Excluding the ground-truth label from the top 5 out of 11 classes is harder than that of 65 and 60 classes.


Table \ref{table_attack} Right shows the SA results. SA is the most difficult because randomly selected class labels often come from significantly different action classes. Although it is might be easier to confuse the model between `deposit' and `grab', it is extremely difficult to do so for `jumping' and `wear-a-shoe'. However, even under such circumstances, SMART is still able to succeed in more than 70\% cases on average, with multiple tests above 96\% and even achieving 100\%.


\subsubsection{Attack Behavioural Analysis.}
\begin{figure}[tb]
  \centering
  \includegraphics[width=0.45\textwidth]{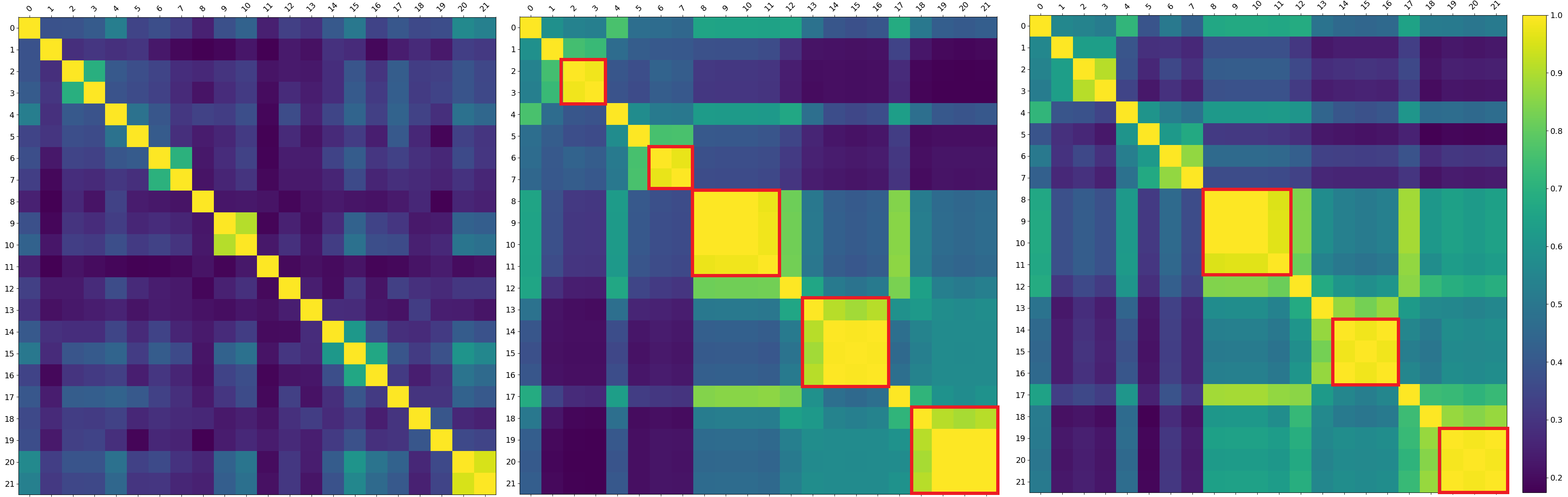}
  \includegraphics[width=0.45\textwidth]{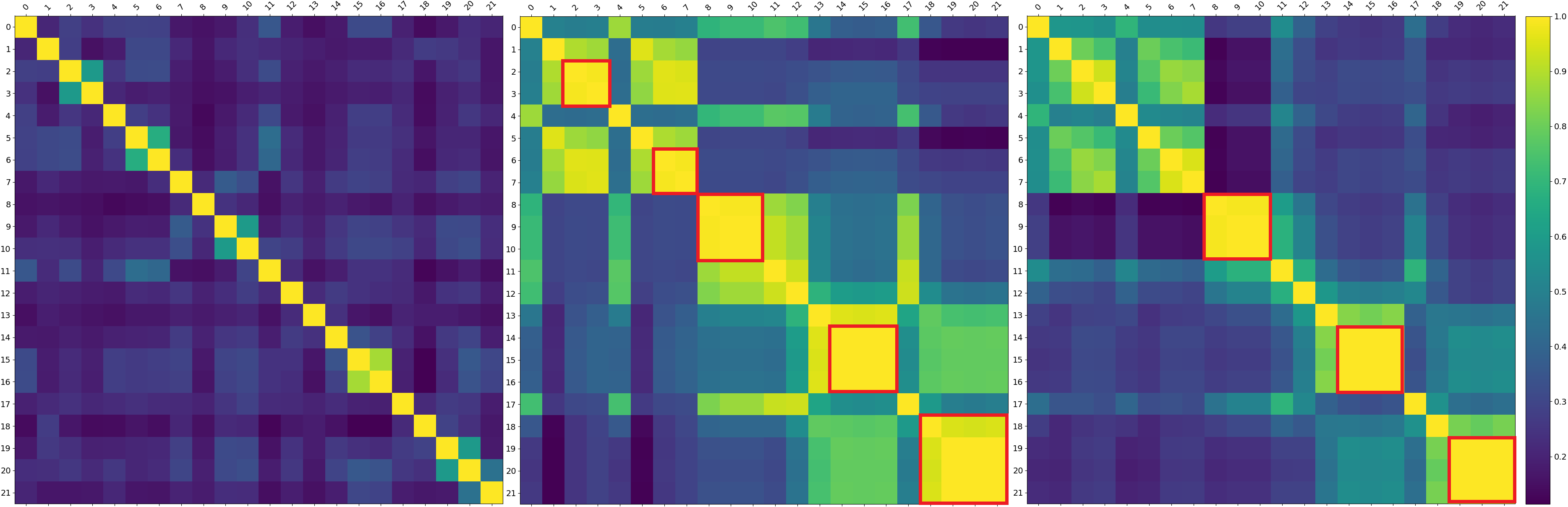}
  \caption{Left: 2S-AGCN on HDM05. Right: DGNN on MHAD. Within each group, displacement-displacement correlations (Left), displacement-speed correlations (Middle) and displacement-acceleration correlations (Right).}
  \label{fig:pertCorr}
\end{figure}

We also analyze the behaviour of SMART by looking at which joint or joint groups are vulnerable. Initially, we thought that if some joints tend to be attacked together, the correlation between the displacements of these joints should be high. So we compute the $l_2$ norm of joint displacements after the attack and their Pearson correlations. We show the results of HDM05 and MHAD on 2SAGCN and DGNN respectively using AB in Figure \ref{fig:pertCorr} (the first image within each group). Although some local high correlations between joint 2 and 3, 6 and 7, 9 and 10, 20 and 21 can be found, they are not universal. Please see other results in the supplementary material. Then we tried to find if across-joint correlations are action-dependent. But no universal conclusion was found either.


Finally, the displacement-speed and displacement-acceleration correlations can give a consistent description of SMART, shown in Figure \ref{fig:pertCorr} (the middle and right image within each group). The correlations are computed between the joint displacements and the original velocities and accelerations, respectively. These two correlations reveal the behaviour of SMART: the higher the speed/acceleration is, the more the joint is attacked (shown by the high values along the main diagonal). In addition, they also reveal some consistent across-joint correlations (as shown by red squares). Note that the joints in a red square belong to one part of the body (four limbs and one trunk). Finally, this also suggests that joints with high velocity and acceleration are important features in the target models because these joints are attacked the most.



\subsection{Black-box Attack}
\label{exp_black}
In the black-box setting, we need a surrogate model to fool the target models. To this end, we use three models: AS-GCN, DGNN and 2s-AGCN, and one dataset: NTU. This is because these models are the latest state-of-the-art methods and their original implementations are available online. In addition, NTU dataset is the one used in all three papers. To test the generalizability of SMART, we in turns take every model as the surrogate model and produce adversarial examples using AB and AB5. Then we use the adversarial examples to attack the other two models. Results are shown in Table \ref{table_bb}.

\begin{table}[tb]
	\centering
	\resizebox{0.5\textwidth}{!}{
		\begin{tabular}{c|ccc}		
			\hline
			Surrogate/Target & DGNN & 2s-AGCN  & AS-GCN              \\ \hline
			DGNN(AB/AB5)                     & n/a  & 90.6(90.99) & 7.24(7.63)        \\
			2s-AGCN(AB/AB5)                      & 98.37(98.46)   & n/a & 98.10(98.96)      \\
			AS-GCN(AB/AB5)                     & 10.90(12.97)   & 91.17(91.99) & n/a      \\
			\hline
		\end{tabular}
	}
	\caption{Success rate of black-box attack.}
	\label{table_bb}
\end{table}

Firstly, AB5 results are in general better than AB. We speculate that there are two factors.  First, the predictive class distribution of AB5 is likely to be flatter than AB. The flatness improves the transferability because a target model with similar decision boundaries will also produce a similarly flat predictive distribution, and thus is more likely to be fooled. Besides, since the ground-truth label is pushed away from the top 5 classes in the surrogate model, it is also likely to be far away from the top in the target model.

We also notice that the transferability is not universally successful. DGNN and AS-GCN cannot easily fool one another. Meanwhile, 2S-AGCN can fool and be fooled by both of them. Since the transferability can be described by distances between decision boundaries \cite{Tramr2017TheSO}, our speculation is that 2S-AGCN's boundary structure overlaps with both DGNN and AS-GCN significantly but the other two overlap little. The theoretic reason is hard to identify, as the formal analysis on transferability has just emerged on static data \cite{Tramr2017TheSO,Zhao_AAAI_2019}. The theoretic analysis of time-series data is beyond the scope of this paper and is therefore left for future work.

\subsection{Perceptual Study}
\label{exp_study}
Imperceptibility is a requirement for adversarial attack. All the success rate shown above would have been meaningless if the attack were noticeable to humans. In other words, success rate alone is not an appropriate evaluation metric. When it comes to imperceptibility, qualitative visual comparisons can be used on image-based attack, but rigorous perceptual studies are needed for complex data \cite{Xiao:2019:MeshAdv}, as high success rate can always be achieved by sacrificing the imperceptibility. This is especially the case for motions. Consequently, the necessity of perceptual studies also restricts us from using certain datasets (e.g. Kinetics \cite{DBLP:journals/corr/KayCSZHVVGBNSZ17}) as baseline due to excessive jittering and tracking losses.

To investigate the imperceptibility, we conducted three user studies (Deceitfulness, Naturalness and Indistinguishability). Since our sample space is huge (5 models $\times$ 3 datasets $\times$ 3 attacking strategies), we chose the most representative model and data. We chose 2S-AGCN as our model because it is one of the latest state-of-the-art methods, with AB as the attacking strategy. We use HDM05 and MHAD datasets. NTU dataset is only used in visual evaluation, not perceptual study due to motion jittering in the original data (see the video for details). In total, we recruited 37 subjects (age between 18 and 37).

{\bf Deceitfulness}. In each user study, we randomly chose 100 motions with the ground-truth and attacked label for 100 trials. In each trial, the video was played for 6 seconds and then the user was asked to choose which label best describes the motion with no time limit. This is to test whether SMART visually changes the semantics of the motion.

{\bf Naturalness}. Since unnatural motions can be easily identified as the result of attack, we performed ablation tests on different loss term combinations. We designed four settings: l2, l2-acc, l2-bone, SMART. l2 is where only the $l_2$ error of joint locations is used, l2-acc is l2 plus the acceleration profile loss, l2-bone is l2 plus the bone-length loss and SMART is our proposed perceptual loss. We first show qualitative comparisons in Figure \ref{fig:ablation}. Video comparisons are available in the supplementary video. Visually, SMART is the best. Even from static postures, one can easily see the artifacts caused by joint displacements. The spine joints are the most obvious. The joint displacements causes unnatural zig-zag bending in l2, l2-acc and l2-bone. 

\begin{figure*}
        \centering
        \includegraphics[width=\linewidth]{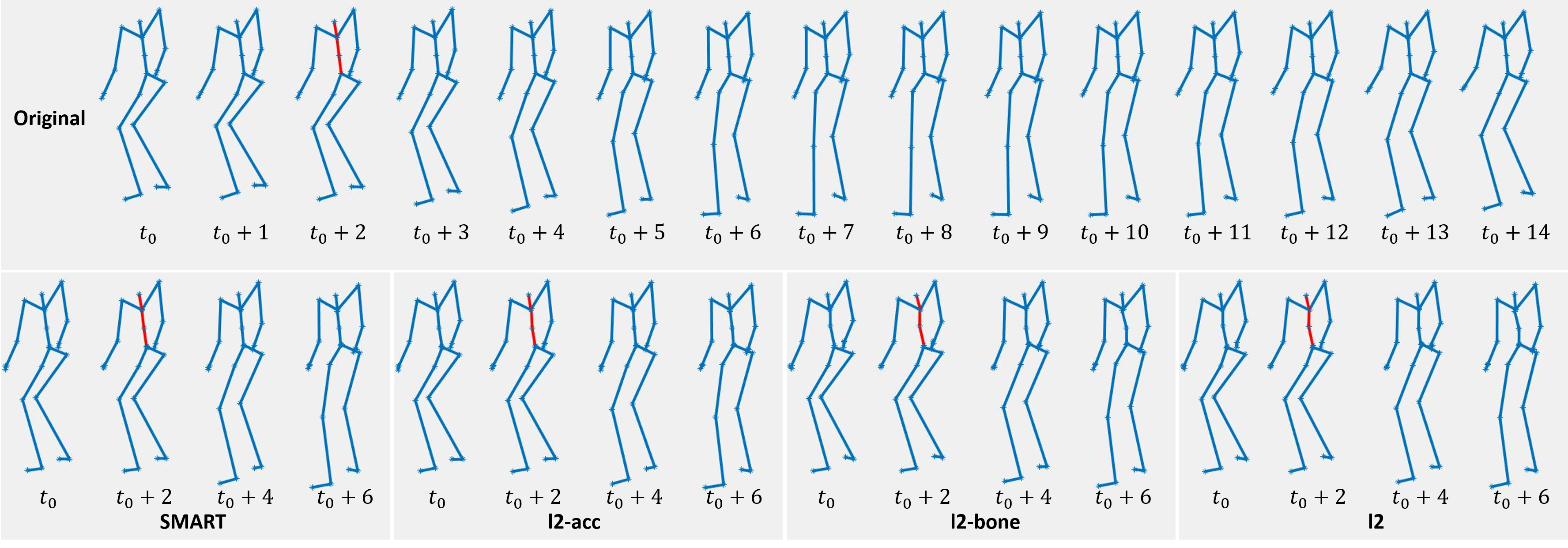}
        
        
  \caption{Visual ablation test between different loss term combinations. Highlighted spine areas in the same frame show key visual differences.}
  \label{fig:ablation}
\end{figure*}

Next, we conducted user studies. In each study, we randomly selected 50 motions. For each motion, we made two trials. The first includes one attacked motion by SMART and one randomly selected from l2, l2-acc and l2-bone. The second includes two motions randomly drawn from l2, l2-acc and l2-bone. The first trial aims to evaluate our results against other alternatives and the second gives insights to the impact of different perceptual loss terms. In each of the 100 trials, two motions were played together for 6 seconds twice, and then the user was asked to choose which motion looks more natural or cannot tell the difference, with no time limit.

{\bf Indistinguishability}. In this study, we did a very strict test to see if the users can tell if a motion is changed in any way at all. In each experiment, 100 pairs of motions were randomly selected. In each trial, the left motion is always the original and the user is told so. The right one can be the original ({\bf sensitivity}) or attacked ({\bf perceivability}). We ask if the user can see any visual differences. Each video is played for 6 seconds then the user was asked to choose if the right motion is a changed version of the left, with no time limit. This user study serves two purposes. Perceivability is a direct test on Indistinguishability on the attack while sensitivity is to screen out subjects who tend to give random choices. Most users are able to recognize if two motions are the same (close to 100\% accuracy), but there are a few whose choices are more random. We discard any user data which falls below 80\% accuracy on the sensitivity test.

\subsubsection{Results.}

\begin{figure}[tb]
  \centering
  \includegraphics[width=0.35\textwidth]{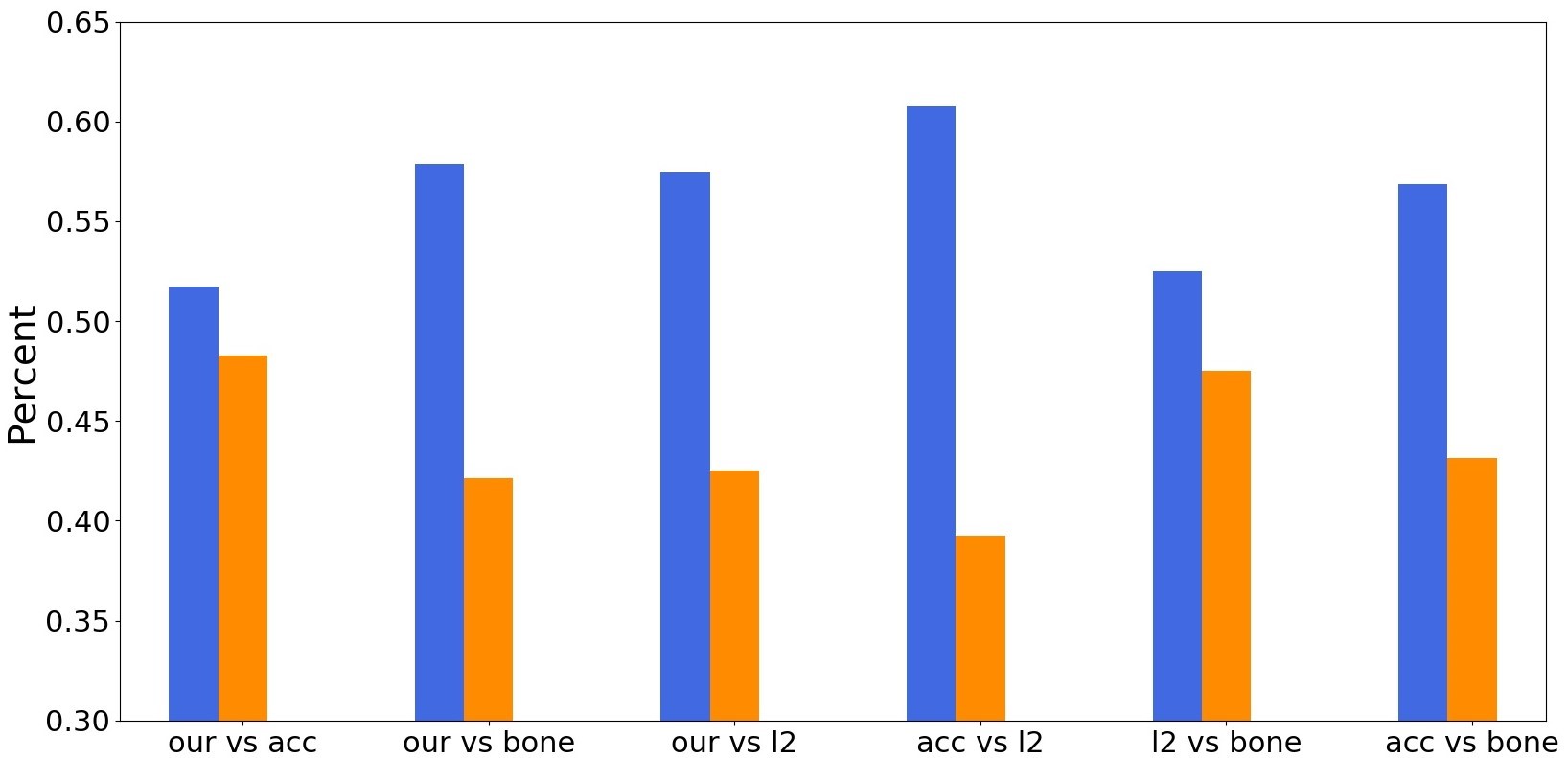}
  \includegraphics[width=0.35\textwidth]{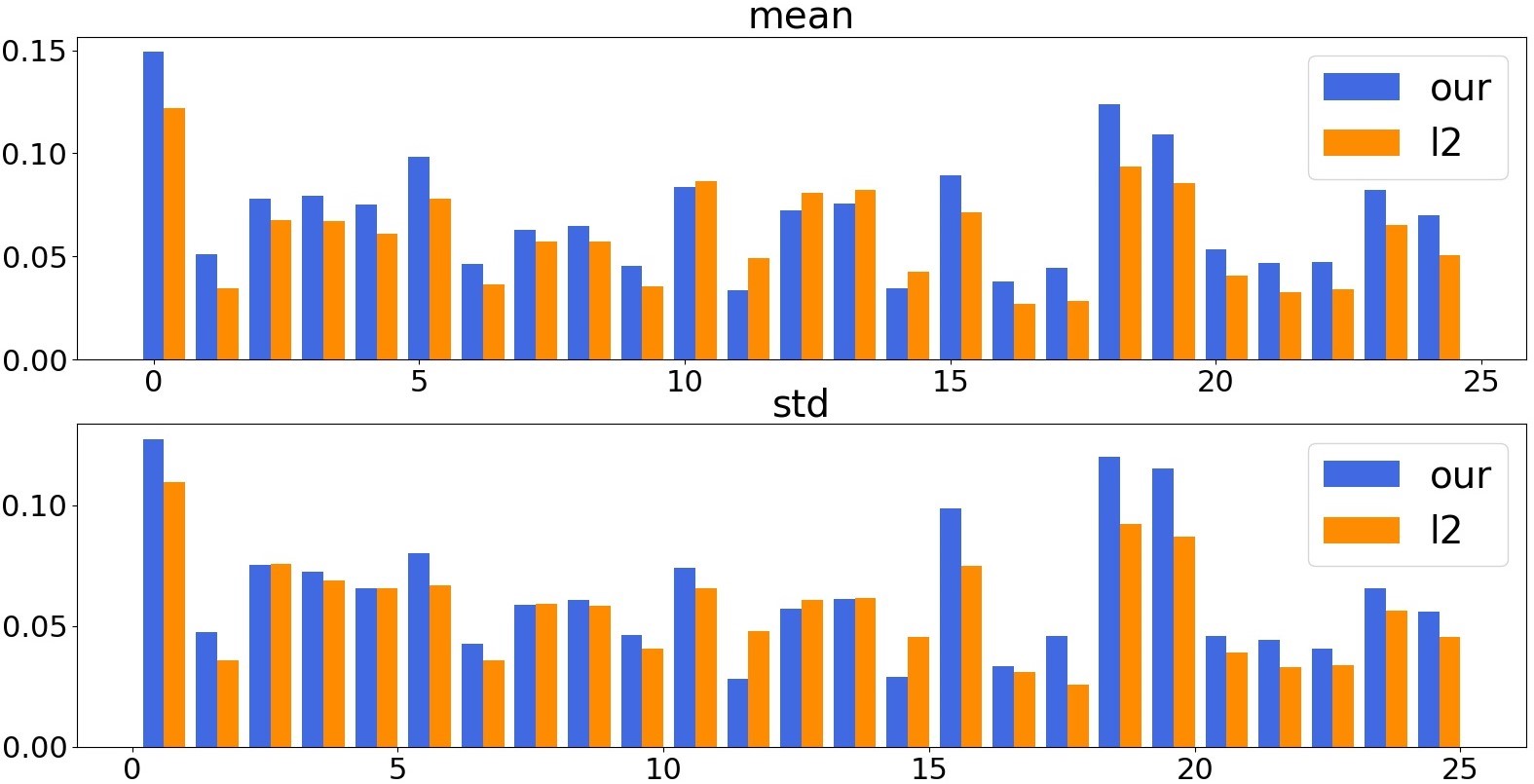}
  \caption{Left: Normalized user preferences on Naturalness. our: SMART. bone: l2-bone. acc: l2-acc. The vertical axis is the percentage of user preference on a particular loss. Right: The mean (Top) and standard deviation (Bottom) of the joint-wise deviations of SMART and l2.}
  \label{fig:userStudy}
\end{figure}

The success rate of {\bf Deceitfulness} is 93.32\% overall, which means that most of the time SMART does not visually change the semantics of the motions. When looking into the success rate on different datasets, SMART achieved 86.77\% on HDM05 and 96.38\% on MHAD. So we looked into in what motions SMART did change the visual semantics. We discovered that the confusion was caused by the motion ambiguity in the original data and labels. For instance, when a `Hopping' motion is attacked into a `Jumping' motion, some users chose `Jumping'. Similar situations occur for `Waving one hand' with `Throwing a ball', `Walking forward' with `Walking LC'. All these motions have small spatial variations and do not distinguish well. Next, Figure \ref{fig:userStudy} Left shows the results of {\bf Naturalness}. Users' preferences over different losses are SMART $>$ l2-acc $>$ l2 $>$ l2-bone. SMART leads to the most natural results as expected.

\textit{Dynamics in Imperceptibility}. To investigate the benefits of exploiting dynamics compared to joint-only perturbation, we did further analysis on SMART-vs-l2 where users chose SMART over l2. We first compute their respective joint-wise deviations from the original motions, shown in Figure \ref{fig:userStudy} Right. In general, the perturbations of SMART are in general higher than l2 and have larger standard deviations. However, the users still chose SMART over l2. It indicates that with proper exploitation of dynamics, larger joint deviations can generate even more desirable results. This is significantly different from the static data (e.g. images), where it is believed that $l$-norms are tightly tied to imperceptibility~\cite{DBLP:journals/corr/abs-1907-10823} (under a $\epsilon$-ball constraints).



Finally, we did the {\bf Indistinguishability} test. The final results are 81.9\% on average, 80.83\% on HDM05 and 83.97\% on MHAD. Note that this is a side-by-side comparison and thus is very harsh on SMART. The users were asked to find any visual differences they could. To avoid situations where motions are too fast to spot any differences (e.g. kicking and jumping motions), we also played the motions three times slower than the original.  Even under such harsh tests, humans still cannot spot any difference most of the time.

\subsection{Comparison}
To our knowledge, the problem we are addressing has not been tackled before, except in the contemporaneous work of Liu et al.~\cite{Liu:2019:AAS} (on arXiv only). Qualitatively, while both methods can achieve high success rate, CIASA \cite{Liu:2019:AAS} generates \textit{plausible} adversarial samples while SMART aims for \textit{natural} and \textit{visually indistinguishable} samples. This is because CIASA uses Generative Adversarial Networks to train a discriminator to judge if every attacked frame is from the real pose distribution without the dynamics, and thus cannot guarantee the imperceptibility. SMART on the other hand is optimized to pass harsh perceptual studies to achieve the imperceptibility. Quantitatively, since both methods attack ST-GCN \cite{Yan_AAAI_2018} on NTU, we directly compare the success rate. While CIASA's best success rate (among alternative settings) is \textit{99.8\%}, SMART achieves \textit{100\%}. 

Although SMART performs better numerically, we would like to point out that numerical comparison alone is not entirely rigorous, because the success rate and perceptual studies need to be coupled in evaluation, as high success rate can always be achieved by sacrificing the imperceptibility and vice versa. However, this unfortunately creates difficulties in conducting further numerical comparisons between the two methods, because the code of CIASA is not available and the paper does not provide enough information for implementation. Moreover, the meta-parameters of CIASA would require hand-crafting for a fair evaluation and it is unclear what the best setting is.


\section{Discussion}
Imperceptibility is vital in adversarial attack. When it comes to skeletal motions, perceptual studies are essential because existing metrics (e.g. $l$-norm) cannot fully reflect perceived realism/naturalness/quality. In addition, it helps us to uncover a unique feature of attacking skeletal motions. Losses solely based on joint location deviations are often overly conservative. It is understandable because they are mainly used for static data and thus are not able to fully utilize the dynamics. Next, forming the joint deviation as a hard constraint~\cite{Liu:2019:AAS} is not the best strategy in our problem. First, a threshold needs to be given and it is unclear how to set it. Second, restricting joint deviation implies that it is solely the most important factor on imperceptibility. Our perceptual study shows that larger joint deviations can be used if the dynamics are exploited properly. Lastly, we could also use joint angles as representations. However, in practice, we find that perturbing joint angles causes jittering and makes the system hard to optimize, similar to \cite{Wang_STRNN_2019}.

\section{Conclusion and Future Work}
In this paper, we proposed a method, SMART, to attack action recognizers based on 3D skeletal motions. Through comprehensive qualitative and quantitative evaluations, we show that SMART is \textit{general} across multiple state-of-the-art models on various benchmark datasets. Moreover, SMART is \textit{versatile} as it can delivery both white-box and black-box attacks with multiple attacking strategies. Finally, SMART is \textit{deceitful} verified in extensive perceptual studies. 

In the future, we would like to theoretically investigate why the transferability varies between different models under black-box attack. Although there has been research on static data, dynamic data is still not investigated. It involves testing more target models and developing a method to describe the structures of class boundaries. We will also investigate on what attack is in people's blind regions. Our experiments indicate that there might be a blind-region sub-space of motions where changes are not perceivable to humans. Adversarial attacks in this sub-space would lead to better results regarding human perception.
%
%
\bibliographystyle{splncs04}
\bibliography{egbib}
\end{document}